\documentclass[sigconf]{acmart}

\usepackage[utf8]{inputenc} 
\usepackage[T1]{fontenc}    
\usepackage{hyperref}       
\usepackage{url}            
\usepackage{booktabs}       
\usepackage{amsfonts}       
\usepackage{nicefrac}       
\usepackage{microtype}      
\usepackage{xcolor}         
\usepackage{todonotes}
\usepackage{caption}
\usepackage{wrapfig}
\usepackage[shortlabels]{enumitem}

\usepackage{amsmath}
\usepackage{mathtools}
\usepackage{amsthm}
\usepackage{caption}
\usepackage{subcaption}
\usepackage[title]{appendix}

\usepackage{tabularx}
\usepackage{multirow}
\usepackage{algorithm}
\usepackage{algorithmic}
\usepackage{wrapfig}

\newcommand{\tref}[1]{Table~\ref{#1}}

\expandafter\def\expandafter\normalsize\expandafter{%
    \normalsize%
    \setlength\abovedisplayskip{2pt}%
    \setlength\belowdisplayskip{8pt}%
    \setlength\abovedisplayshortskip{-8pt}%
    \setlength\belowdisplayshortskip{2pt}%
}

\copyrightyear{2024}
\acmYear{2024}
\setcopyright{rightsretained}
\acmConference[GECCO '24 Companion]{Genetic and Evolutionary Computation
Conference}{July 14--18, 2024}{Melbourne, VIC, Australia}
\acmBooktitle{Genetic and Evolutionary Computation Conference (GECCO '24
Companion), July 14--18, 2024, Melbourne, VIC, Australia}
\acmDOI{10.1145/3638530.3654431}
\acmISBN{979-8-4007-0495-6/24/07}

\begin{document}

\begin{abstract}

Quality Diversity (QD) has shown great success in discovering high-performing, diverse policies for robot skill learning. 
While current benchmarks have led to the development of powerful QD methods, we argue that new paradigms must be developed to facilitate open-ended search and generalizability.
In particular, many methods focus on learning diverse agents that each move to a different $xy$ position in MAP-Elites-style bounded archives. 
Here, we show that such tasks can be accomplished with a single, goal-conditioned policy paired with a classical planner, achieving $O(1)$ space complexity w.r.t. the number of policies and generalization to task variants.
We hypothesize that this approach is successful because it extracts task-invariant structural knowledge by modeling a \textit{relational graph} between adjacent cells in the archive.
We motivate this view with emerging evidence from computational neuroscience and explore connections between QD and models of \textit{cognitive maps} in human and other animal brains.
We conclude with a discussion exploring the relationships between QD and cognitive maps, and propose future research directions inspired by cognitive maps towards future generalizable algorithms capable of truly open-ended search.

\end{abstract}



\title[Quality Diversity for Robot Learning: Limitations and Future Directions]{Quality Diversity for Robot Learning:~\\ Limitations and Future Directions}

\author{Sumeet Batra}
\affiliation{%
  \institution{University of Southern California}
  \streetaddress{idk}
  \city{Los Angeles}
  \country{United States}}
\email{ssbatra@usc.edu}

\author{Bryon Tjanaka}
\affiliation{%
  \institution{University of Southern California}
  \streetaddress{idk}
  \city{Los Angeles}
  \country{United States}}
\email{tjanaka@usc.edu}

\author{Stefanos Nikolaidis}
\affiliation{%
  \institution{University of Southern California}
  \streetaddress{idk}
  \city{Los Angeles}
  \country{United States}}
\email{nikolaid@usc.edu}

\author{Gaurav Sukhatme}
\affiliation{%
  \institution{University of Southern California}
  \streetaddress{idk}
  \city{Los Angeles}
  \country{United States}}
\email{gaurav@usc.edu}

\maketitle

\section{Introduction}
Emergent complexity via open-ended search for diversity has been a long-standing goal of evolutionary computation (EC), with the potential to enable lifelong learning and generalization for embodied, robotic agents. 
It has been argued (e.g. Novelty Search (NS) \cite{novelty_search}) that \textit{artificial complexity} may arise from diversity search similar to the way \textit{biological complexity} arises from evolutionary diversity search.
The foundational method MAP-Elites (ME) \cite{map_elites} planted the seeds for what has come to be known as Quality Diversity (QD), which optimizes for both high-quality \textit{and} diverse solutions.
Despite their deceptive simplicity, these methods are incredibly robust and powerful, enabling real robots to adapt to damage and motor failure \cite{me_adapt_animals}, the ability to discover a diverse range of locomotion gaits \cite{pga-me, ppga}, discover variations of celebrity faces \cite{dqd}, and more. 

However, the simplifying assumptions of ME-based QD methods that make them so powerful are the same assumptions that overly constrain them. 
In particular, MAP-Elites assumes a bounded archive, which prevents infinite exploration in uninteresting directions, but also prevents the possibility of open-ended search. 
In addition, ME-based methods spawn hundreds, if not thousands, of diverse and independent policies that do not exploit the structural knowledge of the task. 
Of particular concern is the application of this approach to the popular QD task of learning locomotion policies that traverse to unique $xy$ locations in maze-like environments.
To explore these environments, QD methods spawn thousands of policies that iteratively reach novel positions in the environment.
This is not computationally efficient, nor will these policies generalize to different mazes. 
In other robotics tasks, such as discovering diverse locomotion gaits where the archive dimension $k$ can be greater than two, the number of policies required to cover the search space increases exponentially with $k$.
Methods such as Centroidal Voronoi Tessellation MAP-Elites (CVT-ME) \cite{cvt-me} prevent this combinatorial explosion by computing a fixed number of Voronoi cells. However, they still must produce an archive of hundreds or even thousands of policies that do not generalize to novel task variations.

Here, we show how a single goal-conditioned policy paired with Dijkstra's single-source shortest-paths planning algorithm 
can solve the maze exploration task, achieve state-of-the-art results on standard QD benchmarks, and generalize to novel mazes in the sense that only the planner needs to re-plan. 
We put forward this method not to argue for using classical planners over contemporary QD methods, but to motivate an investigation into why classical, graph-based planners are so effective and what the community can learn from them to create future learned methods that generalize.
Our key insight is that QD methods explore and map out abstract search spaces similar to "cognitive map" models of the world present in humans and other animals.
Contemporary neuroscience theorizes the existence of these cognitive maps in the hippocampus, which, unlike QD, biological agents use to extract and memorize structural, \textit{reusable} knowledge that allows them to generalize to different task variations \cite{what_is_cog_map, how_to_build_cog_map, humans_organize_knowledge}. 
Interestingly, the hippocampus and memory retrieval have been implicated in predictive tasks \cite{memory_and_prediction}, such as predicting incoming sensory data.
Sensory data that disagree with the prediction (e.g. errors) become learning signals to improve future predictions.
In other words, the literature suggests that memory is crucial in determining \textit{novel} or \textit{surprising} outcomes, which indeed is the role of archives in MAP-Elites-based QD or the unstructured collection in the case of Novelty Search.

Cognitive maps are hypothesized to be \textit{relational graphical models}, where the structural knowledge is used to build a graph of relationships that connect distinct cells to each other. 
We argue that Dijkstra's algorithm for planning over 2D grid-like spatial environments is a special case of this relational graph, where the structural knowledge is realizing that the edges between adjacent cells in the archive are modeling the Euclidean distance between locations in physical space. 
With this in mind, we propose a novel view of Quality Diversity as a \textit{single agent} that learns \textit{diverse skills} by exploring these abstract spaces and extracts their structural, task-invariant knowledge in order to generalize. 
As a novel research direction for the field, we propose investigating the relationships and shortcomings of current methods compared to biological cognitive maps, which we hypothesize will open paths towards algorithms with greater generalizability and ultimately the ability to perform novelty search in an open-ended and biologically plausible way.
We propose several ideas along this direction and discuss how this view can be extended to robotics tasks beyond maze exploration.

\section{Method}
We consider the problem of pairing a hierarchical low-level control policy with a high-level planner to solve exploration and navigation tasks in locomotion environments. 
The low-level controller executes actions to reach intermediate goal states set by the planner. 

The low-level controller is a goal-conditioned parameterized policy $\pi_{\theta}(s, a)$ trained with reinforcement learning (RL) to reach any $xy$-position within a fixed radius.
The state representation $s$ is the one provided by Brax and augmented with the relative $xy$ distance from the agent's center to the goal $p_{rel} = p_g - p_{agent} \in \mathbb{R}^2$, so that the final state representation becomes $s = <p_{rel}, s_{brax}>$. 
We use the CleanRL~\cite{huang2022cleanrl} implementation of PPO \cite{ppo} to train the policy to map states containing goal information $s$ and actions $a$ to reward $r$. 
The reward function is formulated as follows: 
\begin{equation}
    \begin{split}
            r &= c_g * (\delta_{d \geq 1} ||p_{rel}||_2^2 + \delta_{d < 1} ||p_{rel}||_2) + c_a * ||a||_2^2 \\
            & + c_R * (|R_x| + |R_y|) + c_{\omega} * ||\omega||_2 + c_{alive} * r_{alive}
    \end{split}
\end{equation}
Where $d$ is the Euclidean distance to the goal. 
The reward function penalizes distance to the goal $p_{rel}$, control cost $a$, rotations of the agent around the $x$ and $y$ axes $(R_x, R_y)$, and angular velocity $\omega$, with tuneable coefficients $c_g, c_a, c_R, c_{\omega} < 0$. 
Finally, the we reward the agent for every timestep it's "alive" i.e. does not reach a terminal state with $r_{alive}, c_{alive} > 0$.
During training, we randomly and uniformly sample goals within a fixed radius of $r$ meters. 
The list of hyperparameters used during training can be found in the appendix. 

\subsection{Planner}
We view the archive (whose feature descriptors correspond to the $xy$ location of the robot) as a fully connected graph.
Each $i$th cell corresponds to an $xy$ location and is connected to the adjacent cells along the cardinal axes and the diagonals. 
With this formulation, we can explore the archive with any graph-based planner.
Since we are concerned with finding the optimal path to reach every archive cell, we employ Dijkstra's Algorithm. 
Like our traditional QD counterparts and unlike the traditional use case of Dijkstra's for robot planning, we assume no environment information \emph{a priori}. 
Instead, when "visiting" neighboring cells, the algorithm rolls out the goal-conditioned policy from the current cell $c_{cur}$ to each neighboring cell $c_{nbr}$, and only adds $c_{nbr}$ to the shortest paths set if the agent reaches within $\epsilon$ of $c_{nbr}$ before the episode timeout limit $T$.
Full implementation details are available in the appendix.

\section{Experiments}
We compare our approach to the QD-RL algorithms QD-PG~\cite{qdpg}, PGA-ME~\cite{pga-me}, and DCG-ME~\cite{dcg-me} on two tasks in the Brax simulator: AntTrap and AntMaze, where the agent learns to navigate around a trap and through a maze, respectively. 
We modify AntTrap such that the trap spawns much closer to the agent's initial position than in prior work, so that the agent must learn how to escape the trap early on. 
AntMaze is a 3D re-creation of the "hard maze"~\cite{novelty_search}.
We set the episode length to 250 and 3000, and archive resolution to $32 \times 32$ and $40 \times 40$ for AntTrap and AntMaze, respectively. 

Since we do not train a collection of policies and thus do not have training plots of standard QD-metrics, we instead focus on the results of reproducibility experiments for each algorithm, following prior guidelines~\cite{pga-me-replicate}. 
For baselines, we re-evaluate each agent in the archive 50 times.
For our method, we evaluate the agent and planner's ability to reach a cell (goal-state) for every cell in the archive, averaged over 50 environments. 
Finally, we insert the re-evaluated policies into a ``corrected'' archive and report the following ``corrected'' metrics for the archive (\tref{fig:corrected_metrics}):
\textbf{QD-Score} (sum of all objectives in the archive), \textbf{coverage} (percentage of occupied cells), \textbf{best score} (highest undiscounted, cumulative reward achieved by an agent over an episode), and \textbf{descriptor error mean} (DEM) (mean L2 distance between the descriptor achieved by rolling out the policy and the target descriptor). 
Since our method trained on a different reward function than the baselines, we evaluate our policy according to the reward function used by the baselines.

Our method achieves state-of-the-art results in both environments, achieving the best coverage and DEM on both tasks, as well as competitive QD-scores and best-reward scores with baselines. 
\emph{Critically, our method achieves these results by reusing only a single goal-conditioned policy for both tasks.} 
The policy is trained for 100 million environment steps in 6 minutes on a single RTX 3090 GPU. 
Planning with Dijkstra's algorithm for each task takes 30 minutes since the policy must be rolled out $|E|$ times, where $|E|$ is the number of edges in the graph. 
In contrast, the current state-of-the-art QD method DCG-ME required 15 hours to train an archive on AntMaze with an RTX A6000 GPU, due to the evaluation of thousands of policies on a task where the episode length is quite long. 

\begin{figure*}
  \centering
  \setlength{\tabcolsep}{3.4pt}
  \fontsize{6.5}{6.7}\selectfont
  \begin{tabular}{lrrrrrrrr}
  \toprule
    & \multicolumn{4}{c}{AntTrap} & \multicolumn{4}{c}{AntMaze} \\
  \cmidrule(r){2-5}
  \cmidrule(r){6-9}
   & QD-Score ($\times 10^5$) & Cov & Best & MEM & QD-Score ($\times 10^5$) & Cov & Best & MEM    \\
  \midrule
  GCRL + Dijkstra's (ours) & \textbf{4.48} & \textbf{42.93\%} & 1090 & \textbf{0.39} & 29.11 & \textbf{54.33\%} & 12034 & \textbf{0.52}\\
  DCG-ME (Archive) & 3.43 & 39.14\% & 1007 & 3.54 & \textbf{33.91} & 45.77\%  & 12089 & 5.98\\
  DCG-ME (Policy) & 1.43 & 20.41\% & 871 & 16.57 & 6.39 & 10.38\% & 10876 & 13.09 \\
  PGA-ME & 1.57 & 22.11\% & 1007 & 14.72 & 24.78 & 33.28\% & 12089 & 7.52 \\
  QDPG  & 1.61 & 24.41\% & 1007 & 14.45 & 28.97 & 41.91\% & 12089 & 6.37 \\
  \bottomrule
  \end{tabular}
    \caption[Corrected QD metrics: QD-Score, Coverage (Cov), Best Reward (Best), and Descriptor Error Mean (DEM). Results are averaged over 5 seeds for trap and 3 seeds for maze.]{Corrected QD metrics: QD-Score, Coverage (Cov), Best Reward (Best), and Descriptor Error Mean (DEM). Results are averaged over 5 seeds for trap and 3 seeds for maze (due to computational cost arising from the episode length).}
  \label{fig:corrected_metrics}
\end{figure*}


\section{Discussion}

We showed how a simple approach that combines Dijkstra's algorithm with a goal-conditioned policy solves a primary QD benchmark (maze-exploration)
with $O(1)$ space complexity and generalizes to a novel, unseen maze by only re-running the planner.  
The aim of this work is not to propose unilaterally using classical planners with goal-conditioned policies, but to provide a foundation for asking why this method generalizes and what the field can learn from this comparison.
We hypothesize that many robotics tasks contain \textit{structural knowledge} that is present across variants of the task. 
This structure can be modeled as a \textit{graph} of relationships and used to zero-shot generalize at inference time. 
In the specific tasks reported above, one may view Dijkstra's algorithm as being given structural knowledge in the form of an inductive bias -- that is, a vertex corresponding to an $xy$ location is connected to nearby $xy$ locations, where the edge distance is the Euclidean distance between the two $xy$ coordinates. 
Indeed, treating the 2D archive with $xy$ descriptors as a fully connected graph allows us to exploit this structural knowledge with Dijkstra's algorithm and achieve high coverage across maze environments with only a single goal-conditioned policy. 
The question that naturally arises is, \textit{what about tasks where the structural knowledge is non-Euclidean?}
Contemporary research in neuroscience and their emerging connections to machine learning may provide insights into this question.

\subsection{Related Work}
Humans and other animals have a remarkable ability to extract structural knowledge into "cognitive maps" that enable generalization to new tasks and environments.
Cognitive maps are comprised of a variety of spatially specific types of cells, such as place cells \cite{place_cells} that encode for location and grid cells \cite{grid_cells} capable of task-agnostic localization.
This suggests that biological agents learn \textit{factorized representations} of the environment, i.e., the latent variables are independently encoded and \textit{divorced} from sensory data.
Importantly, the same cellular machinery is reused across physical navigation tasks \textit{and} when navigating abstract search spaces \cite{humans_organize_knowledge}. 
Given the reuse of similar mechanisms across search spaces, it has been proposed that \textit{graphical models} may provide a unifying representation of cognitive maps and their ability to navigate and extract structural knowledge in any arbitrary space \cite{how_to_build_cog_map}.
From this perspective, navigation of physical space is a special case of navigating \textit{any} search space, where the relationship between nodes is the Euclidean distance.

Several parallels can be drawn between MAP-Elites-based QD and cognitive maps. 
For example, a MAP-Elites archive can represent any search space, including a geometric one when the feature descriptors are the $xy$ location, which is the structure that our method exploits to generalize.
More generally, MAP-Elites-based QD and cognitive maps play similar roles in that they enable \textit{structured memorization and retrieval of information} for future prediction, or for judging the novelty of new experiences.
However, cognitive maps have an advantage in that they are the product of optimization processes over evolutionary timescales, resulting in efficient, reusable representations. 
Given the similarity between QD and cognitive maps, we ask whether insights from cognitive maps can lead to more efficient and generalizable QD algorithms.

We identify two key components that may help: learning a factorized representation of the latent, task-invariant structure, and using this to construct a relational graph connecting different skills together.
To recap, factorization refers to the separation between observed sensory data and latent structure, and a decomposition of the latent structure into independent variables.
For example, in the task of producing Elon Musk images of varying age and hair length \cite{dqd}, the sensory observation would be the image of Elon Musk, and the latent variables are the descriptors themselves (age and hair length). 
The implication is that cognitive maps enable agents to reason about changes in \textit{any} celebrity face w.r.t. these variables by learning the latent variables \textit{independent} of sensory data \cite{humans_organize_knowledge}.

\vspace{-5pt}
\subsection{Future Directions}
To achieve the type of generalization seen in biological agents, we propose a different view of QD.
Instead of producing thousands of behaviorally diverse agents for singular tasks, we propose training \textit{single agents} 
with neural memory mechanisms similar to those found in the hippocampus, with the goal of downstream task generalization, future prediction, and the discrimination of novel vs nonnovel states. 
To this end, we propose the following research directions inspired by existing literature connecting cognitive map models to contemporary ML methods.

\textbf{Agents with neural memory.} Transformers have been shown to be \textit{implicit graphical models} \cite{tranformers_as_graphs}, where the attention weights are the graph edges between word embeddings, and the metric space is given by the cosine similarity score. 
Additionally, \cite{tem-transformer} shows how a modification to the attention mechanism in the Transformer architecture jointly performs memory retrieval and path-integration similar to the associative memory mechanisms in the hippocampus.
This suggests that certain transformer-based models of long-term memory such as Continuous Hopfield Networks and the TEM-Transformer~\cite{tem-transformer} may be a suitable replacement for MAP-Elites archives as a form of neural, differentiable memory.
Indeed, \cite{QDTransformer} showed promising results when distilling an archive into a Decision Transformer, albeit via distillation of thousands of independent policies without explicit learning of task-invariant structure. 
While further work is required to understand how biological mechanisms of memory are used for task generalization and future prediction, we discuss some promising existing avenues to extract factorized and generalizable representations for a single agent.

\textbf{Learning factorized representations via world models.} 
A promising direction to extracting task-invariant structural knowledge is by learning explicit dynamics models or "world models" that predict future states. 
World models have enabled agents to learn a diverse array of skills and reuse prior knowledge to quickly fine-tune to new tasks within and outside QD \cite{mastering_world_models, scopa}.
However, the ability of these models to generalize to out-of-context tasks and scenarios is limited, and they are prone to distractor features and adversarial attacks \cite{hilton2020understanding, min_attack_rl}.
How to capture the dynamics of the environment completely factorized from the sensory data is an open question, but there is promising research. 
\cite{vic_reg}, for example, trains a variational autoencoder (VAE) with explicit constraints on minimizing the covariance between latent variables and maximizing the information content of the latent vector, resulting in a \textit{disentangled representation} i.e. factorization and achieving greater performance on downstream tasks. 
We propose investigating world models with such constraints designed to achieve a separation between the latent structure and the sensory data.  
For example, in the classical QD task of discovering locomotion gaits via descriptors of proportion foot contact times, a world model that can predict how changes in joint angle torques affect cadence for each leg independently can hypothetically predict action sequences to achieve all possible foot contact timings without having seen examples of those gaits. 

\section{Conclusion}

We demonstrate how a goal-conditioned policy paired with a classical planner achieves state-of-the-art results on a popular QD benchmark for robot learning.
This motivates the notion that a wide variety of tasks, especially those relevant to QD and robot learning, contain structural knowledge that artificial agents can exploit using cognitive maps, similar to biological agents in nature. 
We propose a novel view of QD as a single agent with a cognitive-map-style memory module that learns diverse skills and the structural knowledge that ties those skills together. 
We suggest that closing the gap between QD and biological cognitive maps, by creating models capable of learning relational knowledge graphs and factorized representations of latent structure, will enable future QD algorithms to generalize and perform truly open-ended search.

\bibliographystyle{ACM-Reference-Format}
\bibliography{refs}

\end{document}